\documentclass[conference]{IEEEtran}
\input epsf
\usepackage{graphicx}
\usepackage{pgfplotstable}
\usepackage{tikz}
\usepackage[ruled,vlined]{algorithm2e}

\title{ Pragmatic Implementation of
Reinforcement Algorithms For
Path Finding On Raspberry Pi
}
\makeatletter
 \newcommand{\linebreakand}{%
      \end{@IEEEauthorhalign}
      \hfill\mbox{}\par
      \mbox{}\hfill\begin{@IEEEauthorhalign}
    }
\makeatother
  \author{\IEEEauthorblockN{1\textsuperscript{st} Serena Raju}
    \IEEEauthorblockA{\textit{Computer Department} \\
    \textit{Fr. C. Rodrigues Institute of Technology}\\
    Navi Mumbai, India \\
   scifiserena@gmail.com}
    \and
    \IEEEauthorblockN{2\textsuperscript{nd} Sherin Shibu}
    \IEEEauthorblockA{\textit{Computer Department} \\
    \textit{Fr. C. Rodrigues Institute of Technology}\\
    Navi Mumbai, India  \\
     sherin.shibu@fcrit.onmicrosoft.com}
    \and
    \IEEEauthorblockN{3\textsuperscript{rd} Riya Mol Raji}
    \IEEEauthorblockA{\textit{Computer Department} \\
    \textit{Fr. C. Rodrigues Institute of Technology}\\
     Navi Mumbai, India  \\
    riya.mol@fcrit.onmicrosoft.com}
    \linebreakand 
    \IEEEauthorblockN{4\textsuperscript{th} Joel Thomas}
    \IEEEauthorblockA{\textit{Computer Department} \\
    \textit{Fr. C. Rodrigues Institute of Technology}\\
     Navi Mumbai, India \\
    joel@fcrit.onmicrosoft.com }
    }

\begin{document}
\maketitle 

\begin{abstract}
In this paper, pragmatic implementation of an indoor autonomous delivery system that exploits Reinforcement Learning algorithms for path planning and collision avoidance is audited. The proposed system is a cost-efficient approach that is implemented to facilitate a Raspberry Pi controlled four-wheel-drive non-holonomic robot map a grid. This approach computes and navigates the shortest path from a source key point to a destination key point to carry out the desired delivery. Q learning \cite{1} and Deep-Q learning \cite{2} are used to find the optimal path while avoiding collision with static obstacles. This work defines an approach to deploy these two algorithms on a robot.  A novel algorithm to decode an array of directions into accurate movements in a certain action space is also proposed. The procedure followed to dispatch this system with the said requirements is described, ergo presenting our proof of concept for indoor autonomous delivery vehicles. 

\end{abstract}


%
\IEEEpeerreviewmaketitle
\section{Introduction}
With the COVID-19 pandemic, the demand for automated delivery robots has quadrupled globally \cite{3}. Autonomous robotics is a topic of great relevance today, especially coupled with Reinforcement Learning algorithms \cite{6}. A  part of research focuses on different layers of Motion Planning, such as strategic decisions, trajectory planning, and control \cite{7}. One of the applications of such systems is indoor delivery robots, which can be a great asset to institutions such as hotels, restaurants, educational institutions, especially hospitals, and many other facilities where the obstacles (walls, doors, etc) are static and known. Few existing ones are focused mainly on feature-matching techniques \cite{5}. The main aspects of an indoor delivery robot include indoor localization, pathfinding, obstacle avoidance, and finally traversal which have been studied and implemented in this project. Indoor localization refers to keeping track of a robot in an indoor environment. Pathfinding and obstacle avoidance is the ability of the robot to find the shortest route from a source to a destination such that the robot does not encounter any obstacles. A successful traversal is achieved when the robot is fine-tuned according to the environment to minimise errors in transit. 

\section{Design and Architecture}
The autonomous robot for finding the best possible path using Reinforcement Techniques such as Q learning and Deep Q learning was built using several software components and hardware components within four subsystems: laptop, API server, Raspberry Pi, and mechanical components. 

\subsection{Laptop}
The reinforcement learning computation is done on a laptop device where the obstacle course is defined and a suitable path is detected. This is followed by testing with both Q-learning and Deep Q-learning algorithm and the required best possible path is received in an array that is then converted in JSON format. 

\subsection{API Server}
The best possible path in JSON format is hosted online using an API (https://jsonblob.com/). The JSON file is in the form
{"array": ["x1", "x2", "x3" … "xn"]} where x1 refers to the first desired direction, x2 refers to the second desired direction and so on. This array is hosted on the API's domain which can later be extracted by the Raspberry Pi module using the required API call. 

\subsection{Raspberry Pi}
The Raspberry Pi accesses the best path by making an API call. It also contains the code to decode the encoded array containing the best path into the required sequence of discrete directions. This sequence of directions is then translated into the specific PWM command signals that are sent to the motor driver.
\subsection{Mechanical components}
Hardware implementation is done using components including chassis, wheels, four DC motors, four Encoder Disks, two encoder modules, L298N Dual H-Bridge motor driver, four rechargeable batteries (for the motor drivers), a power bank (for the Raspberry Pi) and wires. The PWM signals are given to the L298N motor driver which is given a power supply of 12V using the rechargeable batteries, which based on the information received gives commands to the DC motors to go ahead, back, left, or right. The movement of the DC motors is tracked using the encoders and a PID controller rectifies the PWM signal to minimize the error. If both pairs of DC motors are given a power supply to move towards the same direction, the car will go forward or backwards according to the values given to INPUT 1 and INPUT 2. If the DC motors on the left are given INPUT 1 and INPUT 2 values to rotate and the right motors are given values to halt then the car will turn towards its right. If the DC motors on the right are given INPUT 1 and INPUT 2 values to rotate and the left motors are given values to halt then the car will turn towards its left.
\vspace{10mm}
\begin{figure}[!h]
	\center
	\includegraphics[scale=0.5]{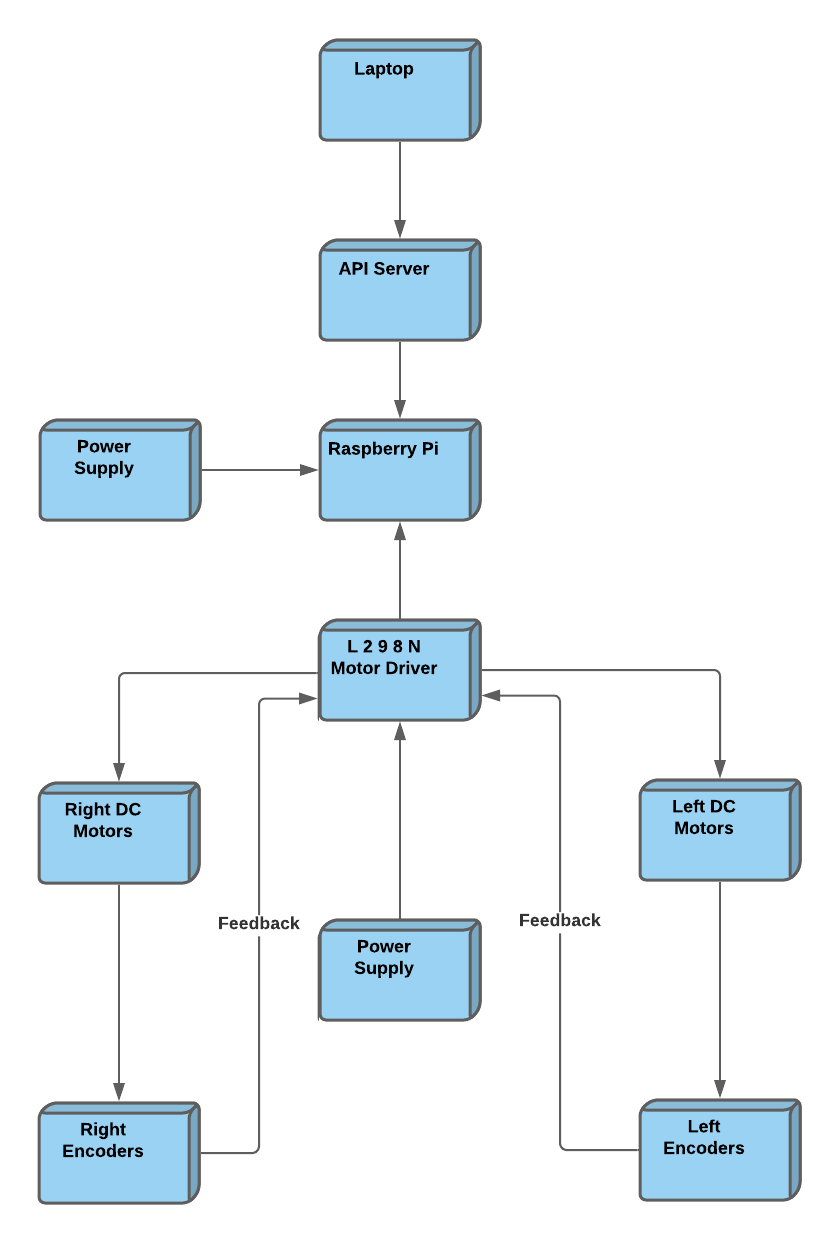} \caption{Block Diagram} \label{f4.1}
\end{figure}

\section{METHODOLOGY OF IMPLEMENTATION}
The proposed system navigates through a known environment beginning from the starting point to the destination point, avoiding all the static obstacles that come along the path. The implementation involves the following elements:

\subsection{User Interface}
The graphical user interface is controlled by the user wherein the essential information such as the number of rows of the grid, the number of columns of the grid, source point, destination point and obstacle points is entered. Separate interfaces for Q-learning and Deep Q-learning models were built.

\subsection{Obtain Optimum Path Using Q-learning Algorithm}
Q-learning is a value-based, off-policy, model-free reinforcement learning algorithm \cite{1} that updates the value function based on an equation. The agent learns from the historical data available for path planning and obstacle avoidance. The most important component of Q-learning is the Q- table which contains Q-values. The Q-table is of the size  $m\times n$  where m is the state size and n is the action size. The action size is fixed with four actions: left, up, right or down. However, the state size will increase exponentially when the size of the matrix changes. The first step is to enter the number of rows and columns and then click the 'Generate Maze' button to create the grid map. The user also enters the starting point, target point and the obstacle points in between after which the agent is trained with the Q-learning algorithm using these initial values.  Successful learning depends on an effective reward function and an efficient selection strategy for obtaining optimal action. The agent selects the action with the highest Q-value among n possible actions when moving from the current state to the next state. The Q-table is constantly updated as iteration increases and the reward is calculated from the agent action. The action and state are updated using the Bellman equation based on the current state and reward. This step is repeated until the final Q-table is obtained and thereby the best path is retrieved from those values. \cite{9}

\subsection{Obtain Optimum Path Using Deep Q-learning Algorithm}
Deep Q-learning overcomes the drawbacks that occur in the case of Q-learning algorithms regarding space and time i.e. in Q-learning the amount of memory and time required to save and update the table increases with an increase in the number of states. This algorithm replaces the regular Q-table with a neural network, namely, the main network and target network wherein for every n step the weights from the main network are copied to the target network \cite{10}. For training the agent on Deep Q-learning the user enters all required values similar to those entered for training on Q-learning. The network has a replay buffer that stores the past experiences after which the Q-network is updated using a subset of these experiences. The main and target model maps the state that is given as input to the output i.e. all possible actions of Q-values are generated. The best-known action of a state is the action that has the largest predicted Q-value.
\subsection{Accessing The Best Path}
The training and testing are done on the laptop device. The optimal path is generated by the above algorithms in the form of an array of numerical values. This array is converted into a JSON file format and hosted on the server using an API. The server is the second subsystem of the proposed system in which the JSON file is updated frequently with the best path until the robot reaches the endpoint. The API is accessed by the Raspberry Pi to determine the next best action.
\subsection{Raspberry Pi Integration and Deployment}
 The Raspberry Pi and the hardware components form the third subsystem of the proposed system. The hardware components required are Raspberry Pi 3 Model B v.1.2, L298N Motor Driver, Chassis Kit with 4 DC motors, Optical Encoder Module, Optical Encoder Disk, DC motors, Male and Female connectors, Power Bank, Switch for the Battery. These components are connected based on the circuit diagram given in Fig. 2.
 \begin{figure}[!h]
	\center
	\includegraphics[scale=0.3]{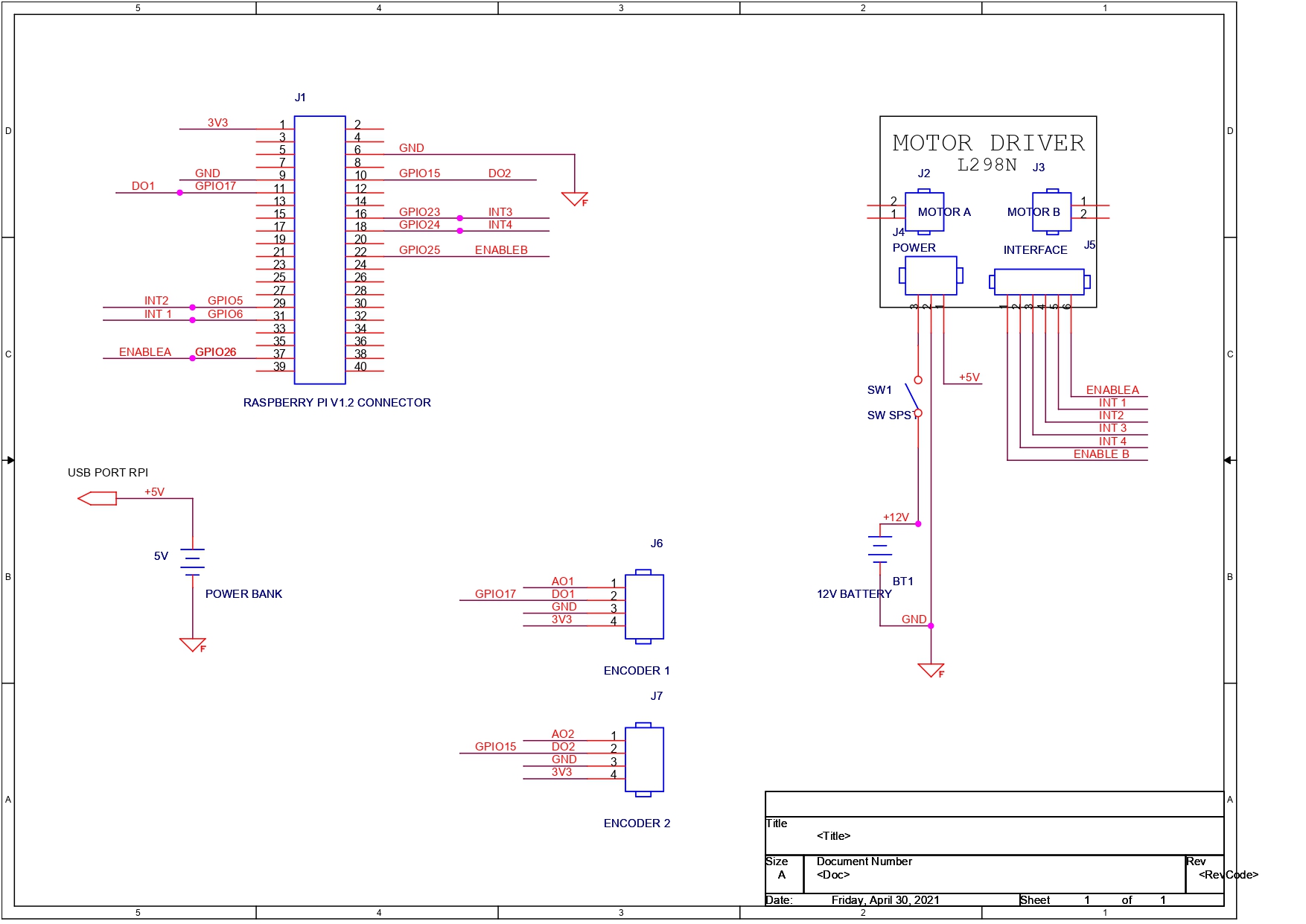} \caption{ Circuit diagram of the robot.} \label{f4.1}
\end{figure}

The array extracted by Raspberry Pi from the API server contains numerical values which indicate directions. The direction and their corresponding numerical values are as follows:\\
LEFT: 0\\
UP:1\\
RIGHT: 2\\
DOWN: 3\\
With these numerical values, the Raspberry Pi sends high/low signals to the motor driver.\\
\begin{table}[!h]
\centering
\begin{tabular}{|r|r|r|}
\hline
Initial direction & Required direction & Movement                  \\ \hline
1                 & 1                  & FORWARD                   \\
1                 & 2                  & \multicolumn{1}{c|}{RIGHT}  \\
1                 & 3                  &                 \multicolumn{1}{c|}{U-TURN}   \\
1                 & 0                  &                       \multicolumn{1}{c|}{LEFT} \\
2                 & 1                  &                   \multicolumn{1}{c|}{LEFT}\\
2                 & 2                  & FORWARD                   \\
2                 & 3                  &             
\multicolumn{1}{c|}{RIGHT} \\
2                 & 0                  &              
\multicolumn{1}{c|}{U-TURN} \\
3                 & 1                  &                 \multicolumn{1}{c|}{U-TURN}\\
3                 & 2                  &                    \multicolumn{1}{c|}{LEFT}\\
3                 & 3                  &                \multicolumn{1}{c|}{FORWARD}\\
3                 & 0                  &                  \multicolumn{1}{c|}{RIGHT}\\
0                 & 1                  &                 \multicolumn{1}{c|}{RIGHT}\\
0                 & 2                  &               \multicolumn{1}{c|}{U-TURN}\\
0                 & 3                  &               \multicolumn{1}{c|}{LEFT }\\
0                 & 0                  & FORWARD               \\ \hline   
\end{tabular}

\end{table}\\
Since the robot receives an encoded array, the array needs to be decoded to orient the robot in the proper direction. For this purpose, instead of keeping all 16 cases as separate cases as shown in the table, a heuristic algorithm was devised to cut down the lines of code. The mathematical relation between the required direction and the encoded numerical values was translated into a simple algorithm. The algorithm is defined in Algorithm 1: Decoding the best path.
\begin{algorithm}[!h]
\SetAlgoLined

\KwIn{Best path array {\bf A} } 
\KwOut{Sequence of required directions from action space}
 \For{$i \leftarrow 0$ \KwTo $len({\bf A})-1$}
 {
  {\bf D} = {\bf A}[i] - {\bf A}[i+1]\;
  \If{{\bf D} = 0}{
   moveForward()\;
   }
   \ElseIf{{\bf D} = -1 or {\bf D} = 3}{
   moveRight()\;
   moveForward()\;
  }
  \ElseIf{{\bf D} = 1 or {\bf D} = -3}{
   moveLeft()\;
   moveForward()\;
  }
  \Else{
  moveLeft()\;
  moveLeft()\;
   moveForward()\;
  }
 }
 \caption{Decoding the best path}
\end{algorithm}
\newline
The motor driver accepts the signals and executes the motion along the course constructed. The encoder detects the motion and sends feedback to the Raspberry Pi to compensate for the error during the motion of the robot. The Raspberry Pi keeps performing the steps of the algorithm until the robot reaches the destination point. The complete activity diagram is given in Fig. 3.
\begin{figure}[!h]
	\center
	\includegraphics[scale=0.35]{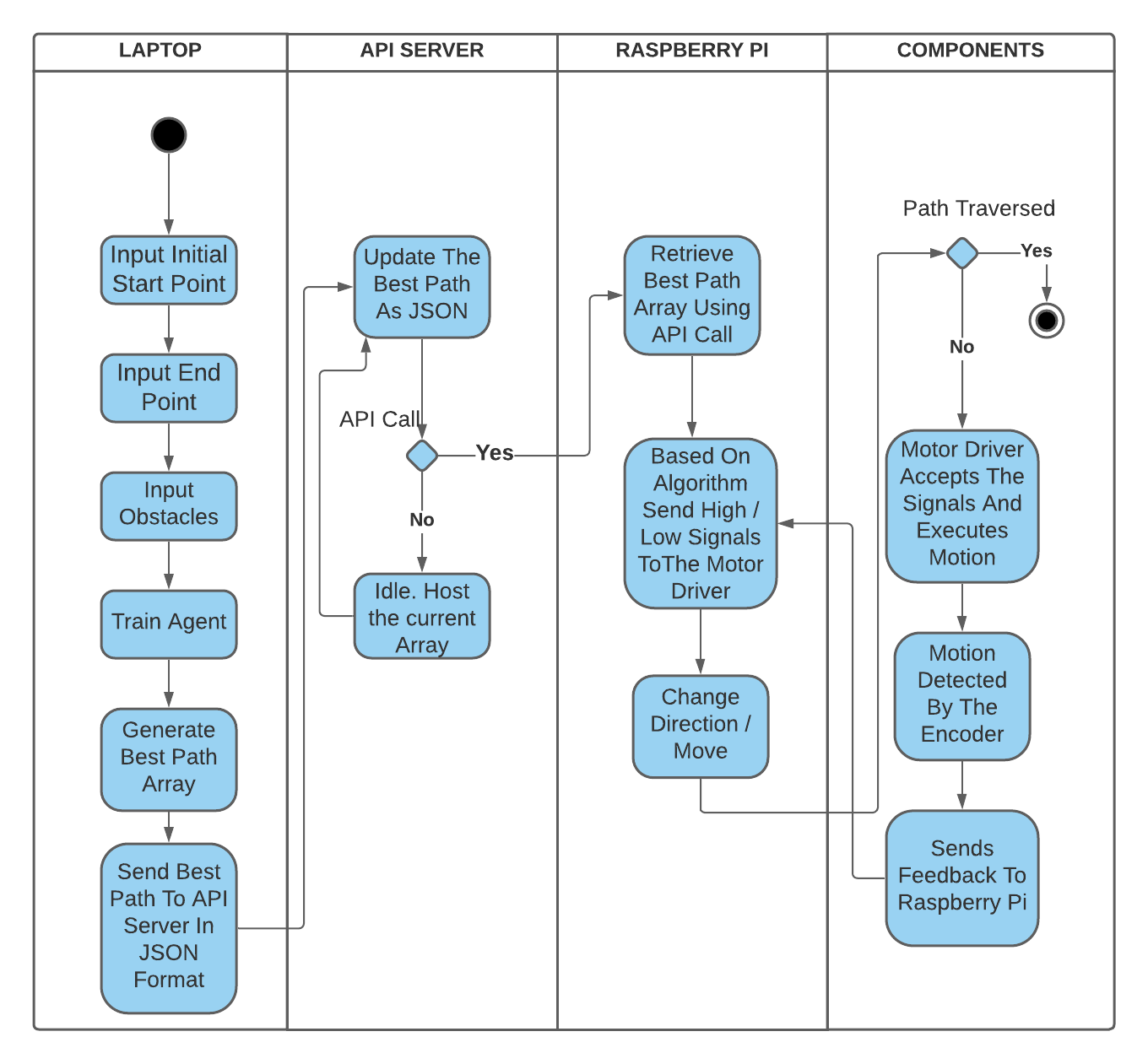} \caption{Complete activity diagram} \label{f4.1}
\end{figure}
\section{RESULTS}
\begin{figure}[!h]
	\center
	\includegraphics[scale=0.2]{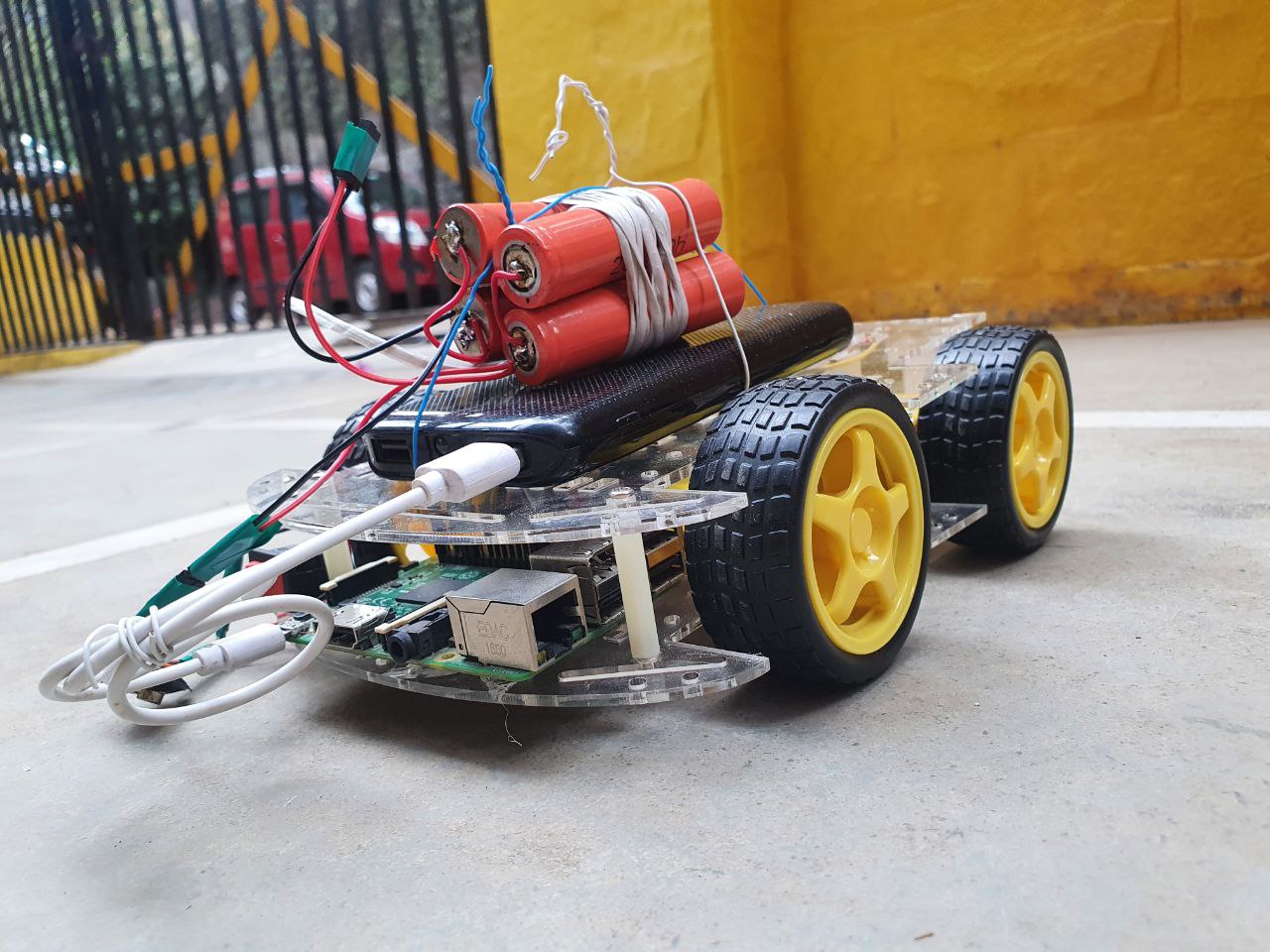} \caption{Assembled Robot} \label{f4.1}
\end{figure}
\begin{figure}[!ht]
    \centering
\pgfplotstableread[col sep=comma,]{try.csv}\datatable
\begin{tikzpicture}
\begin{axis}[
    width=8.5cm,
    height=8cm,
    xtick=data,
    xticklabels from table={\datatable}{Test sample number},
    x tick label style={font=\normalsize, rotate=0},
    legend style={at={(0.65,0.8)},anchor=south east},
    ylabel={Time (minutes)}, xlabel = {Test sample number}]
    
    \addplot [mark=o, blue!80 ] table [x expr=\coordindex, y={Time Taken QL (in minutes)}]{\datatable};
    \addlegendentry{$Q-learning$}
    
    \addplot [mark=o, red!80] table [x expr=\coordindex, y={Time Taken DQL (in minutes)}]{\datatable};
    \addlegendentry{$Deep-Q-learning$}
    
\end{axis}
\end{tikzpicture}
\caption{Velocity comparison of Q-learning and Deep Q-Learning}
\label{fig:my_label}
\end{figure}
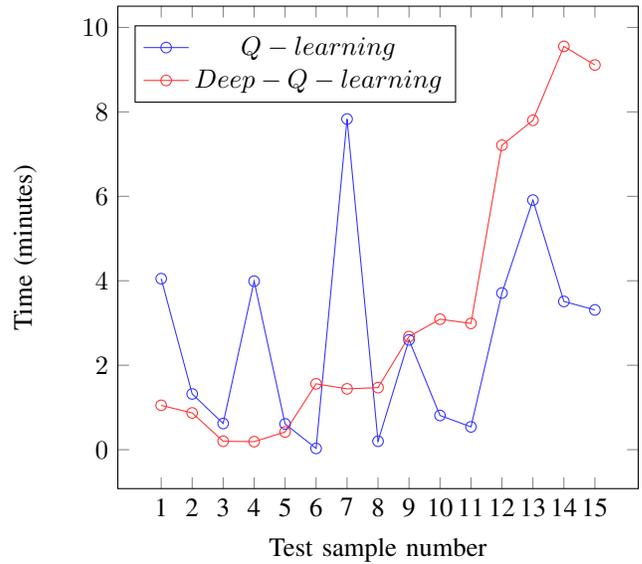

The final assembled robot is shown in Fig. 4. It is a compact four-wheel non-holonomic drive with all the required components attached accurately. Fig \ref{fig:my_label} illustrates a few samples of the time taken by Deep Q Learning and Q Learning training. It is observed that Q Learning performed better than Deep Q Learning in environments where fewer obstacles were present. On the other hand, Deep Q Learning performed better when the number of obstacles was more. 

A few test runs of the robot were conducted. One such test run is described in Fig. 6 and Fig. 7. Fig. 6 is an image of the obstacle course given as input by the user. The orange blocks indicate the obstacles. The top-right pink block is the starting point and the bottom-left pink block is the target point. The black line indicates the path to be followed by the robot which was given as output in the form of an array by the Reinforcement Learning Algorithms.
\begin{figure}[!h]
	\center
	\includegraphics[scale=0.5]{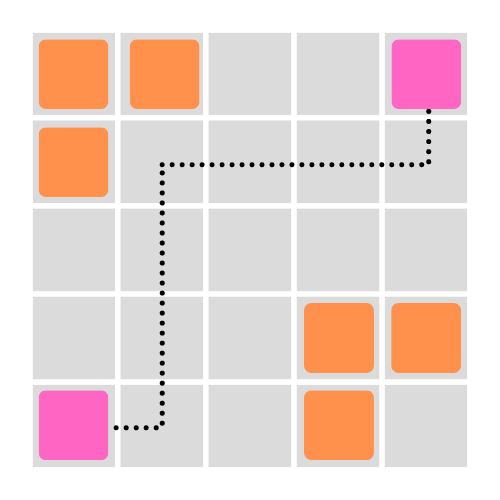} \caption{Sample obstacle course as input} \label{f4.1}
\end{figure}
\begin{figure}[!h]
	\center
	\includegraphics[scale=0.2]{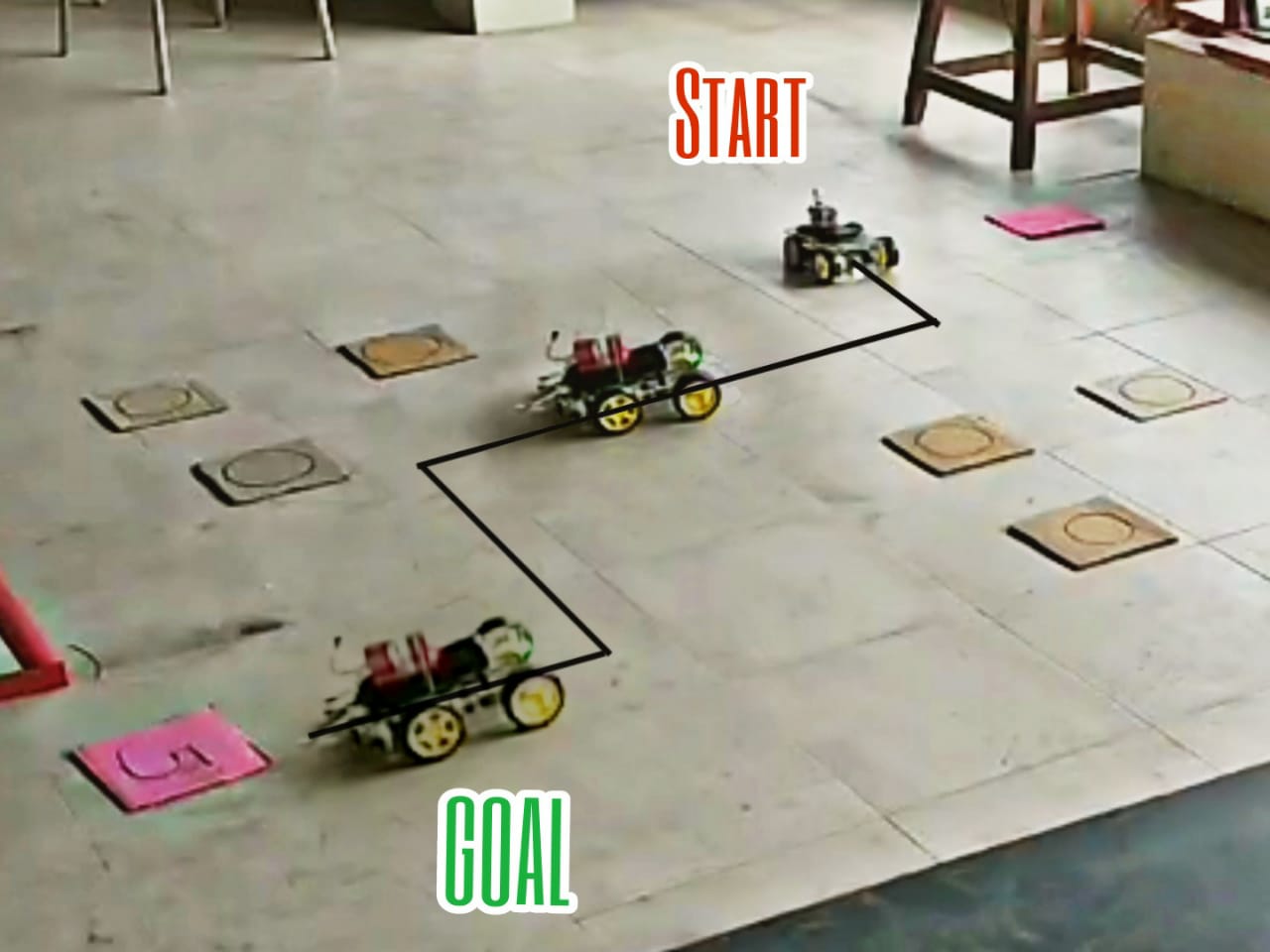} \caption{Edited summary of test run video} \label{res}
\end{figure}
The same obstacle course had been replicated in real-life with the robot placed at the starting point which is shown inFig. 7. The brown cards indicate the obstacles and the pink cards indicate the start and goal. A video was captured with the robot traversing from the starting position to the ending position.
Fig \ref{res} is a multi-exposure image created from the test run's video summarizing the path followed by the robot.

\section{Conclusion}
This project demonstrates an efficient method to build an autonomous indoor delivery robot using a Raspberry Pi. The proposed method effectively eliminates the need for implementing Reinforcement Learning training on the Raspberry Pi itself. This robot was tested on various courses and it can currently traverse through discrete rectangular grids by avoiding static obstacles. In the future this project can be upgraded to reduce the mundane and arduous task of delivering items around a building by including methods to integrate floor plans of buildings as the environment. Implementing this system will benefit all kinds of organizations which require a fast and efficient way to deliver various items within the building.


\section*{Acknowledgment}
The authors would like to extend their sincere gratitude to all the people who made this project successful. We would like to thank our project guide Ms. Smita Dange, Assistant Professor at Fr. C. Rodrigues Institute of Technology, Vashi  for her constant support, guidance and encouragement.




\begin{thebibliography}{1}

\bibitem {1}
Watkins, C.J., Dayan, P. Technical Note: Q-Learning. Machine Learning 8, 279–292 (1992).
https://doi.org/10.1023/A:1022676722315
 \\
\bibitem {2}
Mnih, V., Kavukcuoglu, K., Silver, D. et al. Human-level control through deep reinforcement learning. Nature 518, 529–533 (2015). https://doi.org/10.1038/nature14236
\\
\bibitem {3}
 Edwards, D.(2021, May 19). Demand for delivery robots has ‘quadrupled’ in the past year, says Starship Technologies. Robotics and Automation News. Retrieved from https://roboticsandautomationnews.com/2021/05/19/demand-for-delivery-robots-has-quadrupled-in-past-year-says-starship-technologies/43328/
\\
\bibitem {4}
Sutton, Richard; Barto, Andrew (1998). Reinforcement Learning: An Introduction. MIT Press
\\
\bibitem {5}
Yu Fan Chenand Michael Everett, Miao Liu†, and Jonathan P. Socially aware motion planning with deep reinforcement learning.
\\
\bibitem {6}
Junli Gao †and Weijie Ye, Jing Guo , and Zhongjuan Li. Deep reinforcement learning for indoor mobile robot path planning” sensors 2020, 20(19), 5493;.
\\
\bibitem {7}
Szilárd Aradi ,Survey of Deep Reinforcement Learning for Motion Planning of Autonomous Vehicles .
\\
\bibitem {8}
]Leonid Butyrev1 , Thorsten Edelhäuser ¨ 1 and Christopher Mutschler1,2 Deep Reinforcement Learning for Motion Planning of Mobile Robots.
\\
\bibitem {9}
Fuxiao Tan1(B), Pengfei Yan2, and Xinping Guan3,Deep Reinforcement Learning: From Q-Learning to Deep Q-Learning .
\\
\bibitem {10}
Siyu Zhou, Xin Liu, Yingfu Xu and Jifeng Guo. A Deep Q-network (DQN) Based Path Planning Method for Mobile Robots.




\end{thebibliography}
%

\smallskip
\end{document}